\documentclass[10pt,twocolumn,letterpaper]{article}

\usepackage{iccv}
\usepackage{times}
\usepackage{epsfig}
\usepackage{graphicx}
\usepackage{amsmath}
\usepackage{amssymb}
\usepackage{subfigure}
\usepackage{multirow}
\usepackage{booktabs}
\usepackage[font={small}]{caption}

\usepackage[breaklinks=true,bookmarks=false]{hyperref}

\iccvfinalcopy

\def\iccvPaperID{3592} 

\graphicspath{{./figures/}}

\newcommand{\name} {RF-Action}

\newenvironment{Itemize}%
{\begin{itemize}%
\setlength{\itemsep}{0pt}%
\setlength{\topsep}{0pt}%
\setlength{\partopsep}{0pt}%
\setlength{\parskip}{0pt}}%
{\end{itemize}}
\ificcvfinal\fi
\begin{document}
	\title{Making the Invisible Visible:  Action Recognition Through Walls and Occlusions}
	
	\author{Tianhong Li\thanks{Indicates equal contribution. Ordering determined by inverse alphabetical order.} \quad Lijie Fan$^{\ast}$ \quad Mingmin Zhao \quad Yingcheng Liu \quad Dina Katabi \\ MIT CSAIL}
	
	
	\maketitle

\begin{abstract}
	Understanding people's actions and interactions typically depends on seeing them. Automating the process of action recognition from visual data has been the topic of much research in the computer vision community.  But what if it is too dark, or if the person is occluded or behind a wall?  In this paper, we introduce a neural network model that can detect human actions through walls and occlusions, and in poor lighting conditions.  Our model takes radio frequency (RF) signals as input, generates 3D human skeletons as an intermediate representation, and recognizes actions and interactions of multiple people over time. By translating the input to an intermediate skeleton-based representation,  our model can learn from both vision-based and RF-based datasets, and allow the two tasks to help each other. We show that our model achieves comparable accuracy to vision-based action recognition systems in visible scenarios, yet continues to work accurately when people are not visible, hence addressing scenarios that are beyond the limit of today's vision-based action recognition.

\end{abstract}

\section{Introduction}
Human action recognition is a core task in computer vision. It has broad applications in video games, surveillance, gesture recognition, behavior analysis, etc.  Action recognition is defined as detecting and classifying human actions from a time series (video frames, human skeleton sequences, etc). Over the past few years, progress in deep learning has fueled advances in action recognition at an amazing speed~\cite{qiu2017learning,yue2015beyond,wang2016temporal,shahroudy2016ntu,zhu2016co,fan2018end, huang2018toward,du2015skeleton, fan2019controllable,li2018co,huang2017multi,ke2017new}. Nonetheless, camera-based approaches are intrinsically limited by occlusions -- i.e., the subjects have to be visible to recognize their actions. Previous works mitigate this problem by changing camera viewpoint or interpolating frames over time. Such approaches, however, fail when the camera is fixed or the person is fully occluded for a relatively long period, e.g., the person walks into another room.

Intrinsically, cameras suffer from the same limitation we, humans, suffer from: our eyes sense only visible light and hence cannot see through walls and occlusions. Yet visible light is just one end of the frequency spectrum. Radio signals in the WiFi frequencies can traverse walls and occlusions. Further, they reflect off the human body. If one can interpret such radio reflections, one can perform action recognition through walls and occlusions. Indeed, 
some research on wireless systems  has attempted to leverage this property for action recognition \cite{tian2018rf,yang2011activity,kasteren2010activity,abdelnasser2015wigest,wang2014eyes}. However, existing radio-based action recognition systems lag significantly behind vision-based systems. They are limited to a few actions (2 to 10),  poorly generalize to new environments or people unseen during training, and cannot deal with multi-person actions (see section \ref{sec:related} for details).

\begin{figure}[t]
    \centering
    \includegraphics[width=1.0\linewidth]{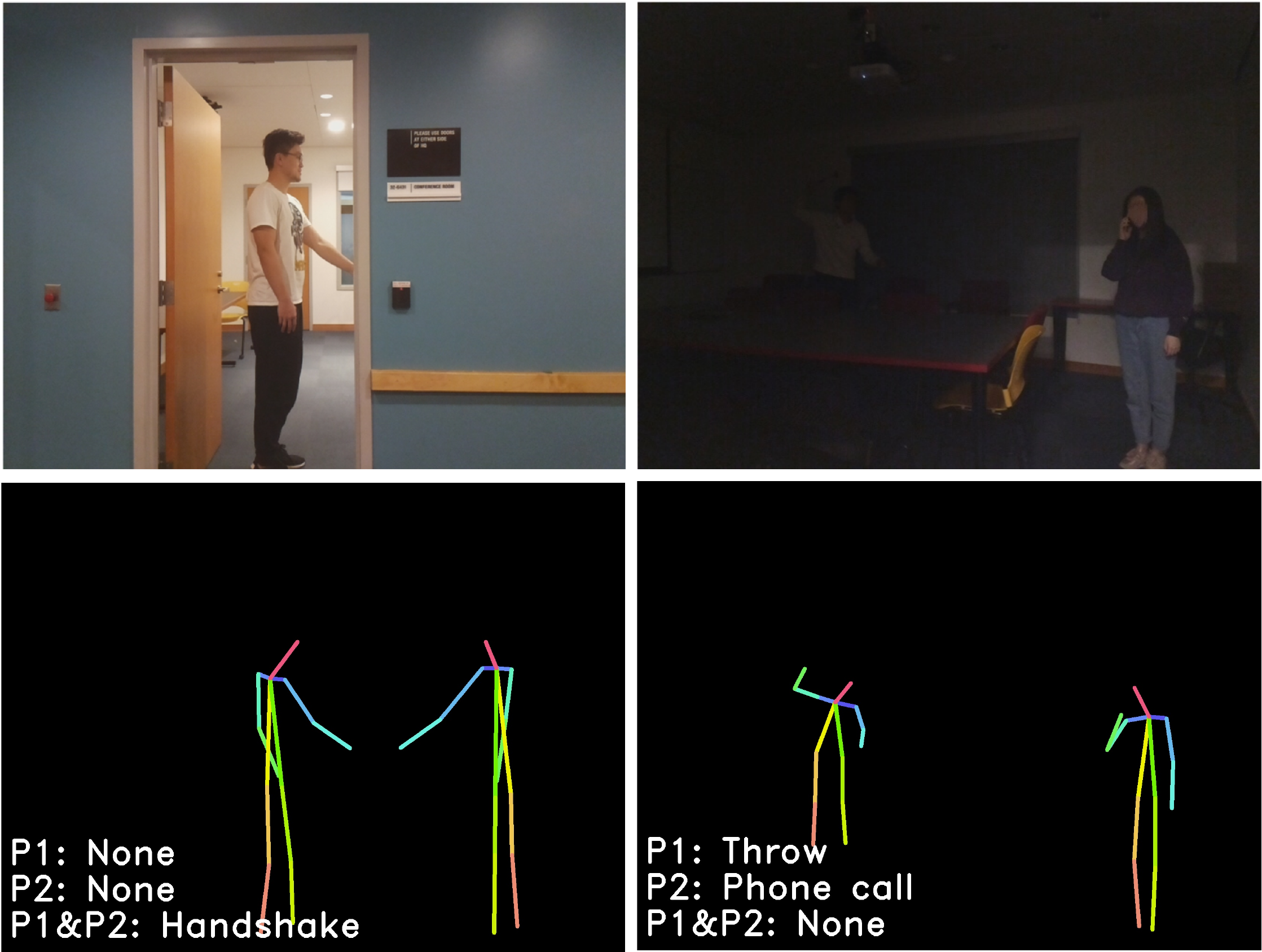}

\caption{\footnotesize{The figure shows two test cases of our system. On the left, two people are shaking hands, while one of them is behind the wall.  On the right, a person is hiding the dark and throwing an object at another person who is making a phone call. The bottom row shows both the skeletal representation generated by our model and the action prediction. }}    \label{fig:teaser}
\vspace{-10pt}
\end{figure}

In this paper, we aim to bridge the two worlds. We introduce, \name, an end-to-end deep neural network that recognizes human actions from wireless signals. It achieves performance comparable to vision-based systems, but can work through walls and occlusions and is insensitive to lighting conditions. Figure \ref{fig:teaser} shows \name's performance in two scenarios. On the left, two people are shaking hands, yet one of them is occluded.  Vision-based systems would fail in recognizing the action, whereas \name\ easily classifies it as handshaking. On the right, one person is making a phone call while another person is about to throw an object at her. Due to poor lighting, this latter person is almost invisible to a vision-based system. In contrast, \name\ recognizes both actions correctly. 

\name\ is based on a multimodal design that allows it to work with both wireless signals and vision-based datasets. We leverage recent work that showed the feasibility of inferring a human skeleton (i.e., pose) from wireless signals~\cite{zhao2018through, zhao2018rf}, and adopt the skeleton as an intermediate representation suitable for both RF and vision-based systems. Using skeletons as an intermediate representation is advantageous because: (1) it enables the model to train with both RF and vision data, and leverage existing vision-based 3D skeleton datasets such as PKU-MMD and NTU-RGB+D~\cite{liu2017pku, shahroudy2016ntu}; (2) it allows additional supervision on the intermediate skeletons that helps guide the learning process beyond the mere action labels used in past RF-based action recognition systems; and (3) it improves the model's ability to generalize to new environments and people because the skeleton representation is minimally impacted by the environment or the subjects' identities. 

We further augment our model with two innovations that improve its performance: First, skeleton, particularly those generated from RF signals, can have errors and mispredictions.  To deal with this problem, our intermediate representation includes in addition to the skeleton a time-varying confidence score on each joint. We use self-attention to allow the model to attend to different joints over time differently, depending on their confidence scores. 

Second, past models for action recognition generate a single action at any time. However, different people in the scene may be engaged in different actions, as in the scenario on the right in Figure \ref{fig:teaser} where one person is talking on the phone while the other is throwing an object. Our model can tackle such scenarios using a multi-proposal module specifically designed to address this issue. 

To evaluate \name, we collect an action detection dataset from different environments with a wireless device and a multi-camera system. The dataset spans 25 hours and contains 30 individuals performing various single-person and multi-person actions. 
%
Our experiments show that \name\ achieves performance comparable to vision-based systems in visible scenarios, and continues to perform well in the presence of full occlusions. Specifically, \name\  achieves 87.8 mean average precision (mAP)  with no occlusions, and an mAP of 83.0 in through-wall scenarios. Our results also show that multimodal training improves action detection for both the visual and wireless modalities. Training our model with both our RF dataset and the PKU-MMD dataset, we observe a performance increase in the mAP of the test set from 83.3 to 87.8 for the RF dataset (no occlusion), and from 92.9 to 93.3 for the PKU-MMD dataset (cross subjects), which shows the value of using the skeleton as an intermediate common representation. 

\vskip 0.06in \noindent
{\bf Contributions:} 
The paper has the following contributions:

\begin{Itemize}
\item 
It presents the first model for skeleton-based action recognition using radio signals; It further demonstrates that such model can accurately recognize actions and interactions through walls and in extremely poor lighting conditions using solely RF signals (as shown in Figure 1).
\item
The paper proposes ``skeletons" as an intermediate representation for transferring knowledge related to action recognition across modalities, and empirically demonstrate that such knowledge transfer improves performance. 
\item
The paper introduces a new spatio-temporal attention module, which improves skeleton-based action recognition regardless of whether the skeletons are generated from RF or vision-based data.
\item
It also presents a novel multi-proposal module that extends skeleton-based action recognition to detect simultaneous actions and interactions of multiple people.
\end{Itemize}
\section{Related Works}
\label{sec:related}

\noindent {\bf (a)~~Video-Based Action Recognition: }
Recognizing actions from videos has been a hot topic over the past several years. Early methods use hand-crafted features. For instances, image descriptors like HOG and SIFT have been extended to 3D \cite{chen2009mosift,ng2018actionflownet} to extract temporal clues from videos. Also, descriptors like improved Dense Trajectories (iDT) \cite{wang2013action} are specially designed to track motion information in videos. More recent solutions are based on deep learning, and fall into two main categories. The first category extracts motion and appearance features jointly by leveraging 3D convolution networks \cite{carreira2017quo,qiu2017learning}. The second category considers spatial features and temporal features separately by using two-stream neural networks \cite{simonyan2014two, wang2016temporal}. 
 
\vskip  0.06in\noindent {\bf (b)~~Skeleton-Based Action Recognition: }
Skeleton-based action recognition has recently gained much attention \cite{fang2017rmpe,cao2018openpose}. Such an approach has multiple advantages. First, skeletons  provide a robust representation for human dynamics against background noise \cite{li2018co}. Second, skeletons are more succinct in comparison to RGB videos, which reduces computational overhead and allows for smaller models suitable for mobile platforms \cite{ke2017new}.

Prior work on skeleton-based action recognition can be divided to three categories. Early work used Recurrent Neural Networks (RNNs) to model temporal dependencies in skeleton data \cite{du2015hierarchical,shahroudy2016ntu,zhu2016co}. Recently, however, the literature shifted to Convolutional Neural Networks (CNNs) to learn spatio-temporal features and achieved impressive performance \cite{du2015skeleton,li2018co,ke2017new}. Also, some papers represented skeletons as graphs and utilized graph neural network (GNN) for action recognition \cite{yan2018spatial, gao2018generalized}. In our work, we adopt a CNN-based approach, and expand on the Hierarchical Co-occurrence Network (HCN) model \cite{li2018co} by introducing a spatio-temporal attention module to deal with skeletons generated from wireless signals, and a multi-proposal module to enable multiple action predictions at the same time.

\vskip  0.06in\noindent {\bf (c)~~Radio-Based Action Recognition: }
Research in wireless systems has explored action recognition using radio signals, particularly for home applications where privacy concerns may preclude the use of cameras~\cite{wang2014eyes, guo2017novel,pu2013whole,abdelnasser2015wigest}.  These works can be divided into two categories:  The first category is similar to \name\ in that it analyses the radio signals that bounce off people's bodies. They use action labels for supervision, and simple classifiers \cite{wang2014eyes, guo2017novel, pu2013whole, abdelnasser2015wigest}.  They recognize only simple actions such as walking, sitting and running, and a maximum of 10 different actions. Also, they deal only with single person scenarios. The second category relies on a network of sensors. They either deploy different sensors for different actions, (e.g., a sensor on the fridge door can detect eating) \cite{kasteren2010activity, yang2011activity}, or attach a wearable sensor on each body part and recognize a subject's actions based on which body part moves \cite{keally2011pbn}. Such  systems require a significant instrumentation of the environment or the person, which limits their utility and robustness.

\section{Radio Frequency Signals Primer}

\begin{figure}[t]
\centering
\includegraphics[width=0.8\linewidth]{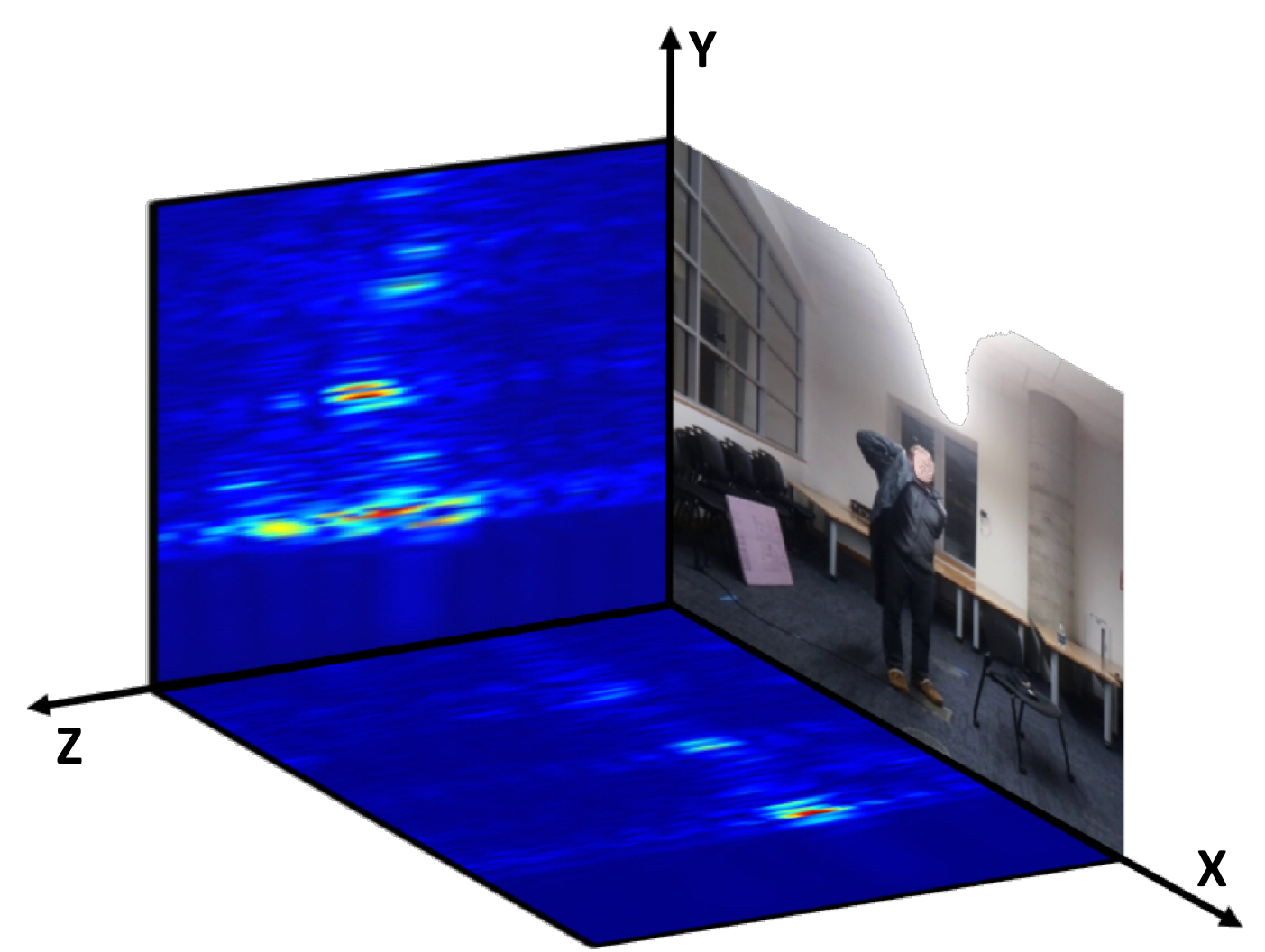}
\caption{\footnotesize{RF heatmaps and an RGB image recorded at the same time.}}	\label{fig:rf-heatmaps}
\end{figure}


We use a type of radio commonly used in past work on RF-based action recognition~\cite{zhao2018rf, lien2016soli, zhang2018latern, chetty2017low, peng2016fmcw, tian2018rf, hsu2019enabling, zhao2016emotion, zhao2017learning, zhao2019through}.  The radio generates a waveform called FMCW 
and operates between 5.4 and 7.2 GHz. The device has two arrays of antennas organized vertically and horizontally.
Thus, our input data takes the form of two-dimensional heatmaps, one from the horizontal array and one from the vertical array. As shown in Figure \ref{fig:rf-heatmaps}, the horizontal heatmap is a projection of the  radio signal on a plane parallel to the ground, whereas the vertical heatmap is a projection of the signal on a plane perpendicular to the ground (red refers to large values while blue refers to small values). 
Intuitively, higher values correspond to higher strength of signal reflections from a location. The radio works at a frame rate of 30 FPS, i.e., it generates 30 pairs of heatmaps per second. 

As apparent in Figure \ref{fig:rf-heatmaps}, RF signals have different properties from visual data, which makes RF-based action recognition a difficult problem. In particular:
\begin{Itemize}
	\item RF signals in the frequencies that traverse walls have  lower spatial resolution than visual data. In our system, the depth resolution is ~10 cm, and the angle resolution is ~10 degrees. Such low resolution makes it hard to distinguish activities such as hand waving and hair brushing.
	\item The human body is specular in the frequency range that traverse walls \cite{beckmann1987scattering}. RF specularity is a physical phenomenon that occurs when the wavelength is larger than the roughness of the surface. In this case, the object acts like a reflector - i.e., a mirror - as opposed to a scatterer. The wavelength of our radio is about 5cm and hence humans act as reflectors. Depending on the orientation of the surface of each limb, the signal may be reflected towards our sensor or away from it. Limbs that reflect the signal away from the radio become invisible for the device. Even if the signals are reflected back to the radio, limbs with a small surface (e.g., hands) reflect less signals and hence are harder to track. 
	\item Though RF signals can go through walls, their attenuation as they traverse a wall is significantly larger than through air. As a result, the signals reflected from a human body is weaker when the person is behind a wall, and hence the accuracy of detecting an action decreases in the presence of walls and occlusions. 
\end{Itemize}

\begin{figure*}[t]
\begin{center}
\includegraphics[width=1.0\linewidth]{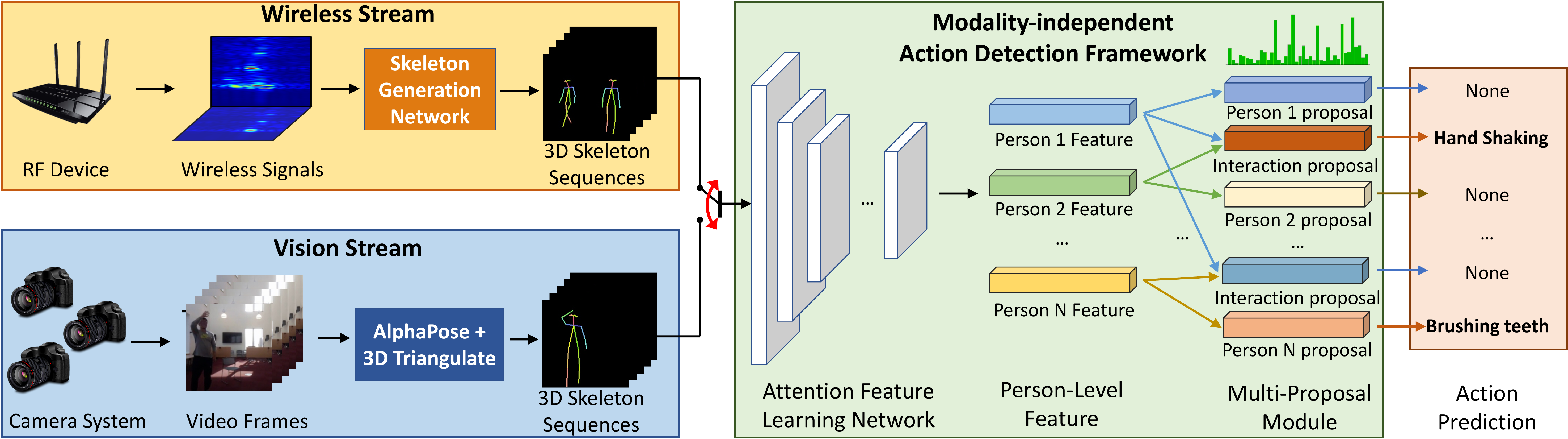}
\end{center}
\vspace{-10pt}
\caption{\footnotesize{\name\ architecture. \name\ detects human actions from wireless signals. It first extracts 3D skeletons for each person from raw wireless signal inputs (yellow box). It then performs action detection and recognition on the extracted skeleton sequences (green box). The Action Detection Framework can also take 3D skeletons generated from visual data as inputs (blue box), which enables training with both RF-generated skeletons and existing skeleton-based action recognition datasets.}}\label{fig:method}
\vspace{-10pt}
\end{figure*}

\section{Method}
\name\ is an end-to-end neural network model that can detect human actions through occlusion and in bad lighting. The model architecture is illustrated in Figure \ref{fig:method}. As shown in the figure, the model takes wireless signals as input, generates 3D human skeletons as an intermediate representation, and recognizes actions and interactions of multiple people over time. 
The figure further shows that \name\ can also take 3D skeletons generated from visual data. This allows \name\ to train with existing skeleton-based action recognition datasets. 
 
In the rest of this section, we will describe how we transform wireless signals to 3D skeleton sequences, and how we infer actions from such skeleton sequences --i.e., the yellow and green boxes in Figure \ref{fig:method}. Transforming visual data from a multi-camera system to 3D skeletons can be done by extracting 2D skeletons from images using an algorithm like AlphaPose and
then triangulating the 2D keypoints to generate 3D skeletons, as commonly done in the literature~\cite{hartley2003multiple,zhao2018rf}.

\subsection{Skeleton-Generation from Wireless Signals}
To generate human skeletons from wireless signals, we adopt the architecture from \cite{zhao2018rf}. Specifically, the skeleton generation network (the orange box in Figure \ref{fig:method}) takes in wireless signals in the form of horizontal and vertical heatmaps shown in Figure \ref{fig:rf-heatmaps}, and generates multi-person 3D skeletons. The input to the network is a 3-second window (90 frames) of the horizontal and vertical heatmaps. The network consists of three modules commonly used for pose/skeleton estimation \cite{zhao2018rf}. First, a feature network comprising spatio-temporal convolutions extracts features from the input RF signals. Then, the extracted features are passed through a region proposal network (RPN) to obtain several proposals for possible skeleton bounding boxes. Finally, the extracted proposals are fed into a 3D pose estimation sub-network to extract 3D skeletons from each of them.


\subsection{Modality-Independent Action Recognition}
As shown in Figure \ref{fig:method}, the Modality-Independent Action Recognition framework uses the 3D skeletons generated from RF signals to perform action detection.

\noindent\textbf{Input: }We first associate the skeletons across time to get multiple skeleton sequences, each from one person. Each skeleton is represented by the 3D coordinates of the keypoints (shoulders, wrists, head, etc.). Due to radio signal properties, different keypoints reflect different amounts of radio signals at different instances of time, leading to varying confidence in the keypoint location (both across time and across keypoints). Thus, we use the skeleton generation network's prediction confidence as another input parameter for each keypoint. Therefore, each skeleton sequence is a matrix of size $4\times T \times N_j$,  where 4 refers to the spatial dimensions plus the confidence, $T$ is the number of frames in a sequence, and $N_j$ corresponds to the number of keypoints in a skeleton.

\noindent\textbf{Model:} Our action detection model (the large green box in Figure \ref{fig:method}) has three modules as follows:
1) An attention-based feature learning network, which extracts high-level spatio-temporal features from each skeleton sequence. 
2) We then pass these features to a multi-proposal module to extract proposals -- i.e., time windows that each corresponds to the beginning and end of an action.  Our multi-proposal module consists of two proposal sub-networks: one to generate proposals for single person actions, and the other for two-people interactions.
3) Finally, we use the generated proposals to crop and resize the corresponding latent features and input each cropped action segment into a classification network. The classification network first refines the temporal proposal by performing a 2-way classification to determine whether this duration contains an action or not. It then predicts the action class of the corresponding action segment.

Next, we describe the attention module and the multi-proposal module in detail.

\subsubsection{Spatio-Temporal Attention Module}\label{sec:attention}
We learn features for action recognition using a spatio-temporal attention-based network. 
Our model builds on the hierarchical co-occurrence network (HCN) \cite{zhu2016co}. HCN uses two streams of convolutions:
a spatial stream that operates on skeleton keypoints, and a temporal stream that operates on changes in the locations of the skeleton's keypoints across time.  HCN concatenates the output of these two streams to extract spatio-temporal features from the input skeleton sequence. It then uses these features to predict human actions.   

\begin{figure}[t]
\centering
\includegraphics[width=0.8\linewidth]{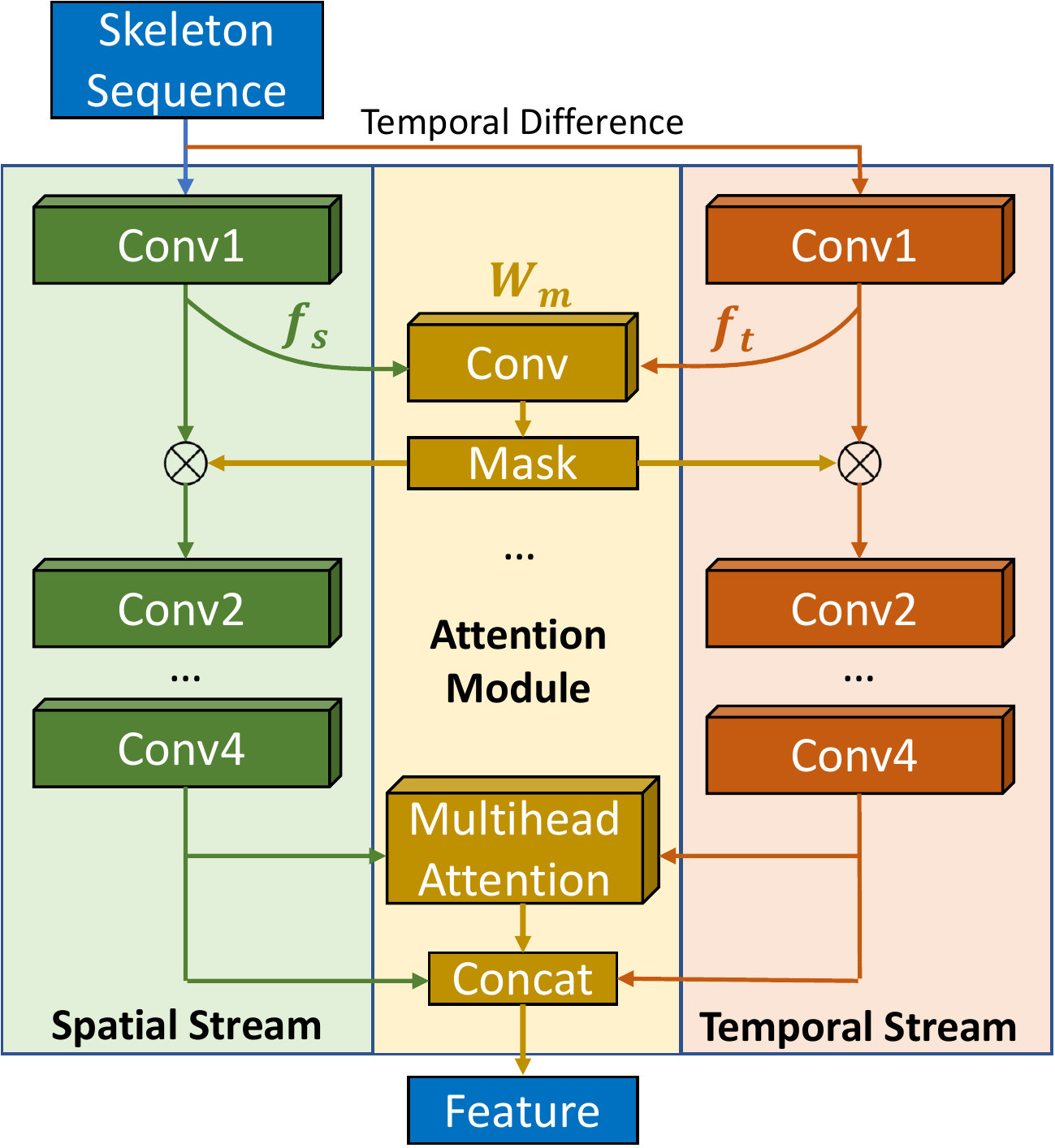}
\caption{\footnotesize{Spatio-temporal attention module. Our proposed attention module (yellow box) learns masks which make the model focus more on body joints with higher prediction confidence. It also uses a multi-headed attention module to help the model attend more on useful time instances.}}    \label{fig:attention}
\vspace{-5pt}
\end{figure}

However, skeletons predicted from wireless signals may not be as accurate as those labeled by humans. Also, different keypoints may have different prediction errors. To make our action detection model focus on body joints with higher prediction confidence, we introduce a spatio-temporal attention module (Figure \ref{fig:attention}). Specifically, we define a learnable mask weight $W_m$, and convolve it with latent spatial features $f_{s}$, and temporal features $f_{t}$ at each step:
$$Mask=Conv(concat(f_{s}, f_{t}), W_m).$$ 

We then apply the $Mask$ on the latent features as shown in Figure \ref{fig:attention}.  In this way, the mask could learn to provide different weights to different joints to get better action recognition performance. We also add a multi-headed attention module \cite{vaswani2017attention} on the time dimension after the feature extraction to learn the attention on different timestamps.

Our proposed attention module helps the model to learn more representative features, since the learnt mask leverages information provided by both the spatial stream and temporal stream, and the multi-headed attention helps the model to attend more on useful time instances. This spatio-temporal attention changes the original HCN design where the spatial and temporal path interact with each other only using late fusion. Our experiments show that the spatio-temporal attention module not only helps increase the action detection accuracy on skeletons predicted from wireless signals, but also helps increase the performance on benchmark visual action recognition datasets. This further shows the proposed attention module helps to combine spatial and temporal representations more effectively, and would lead to better feature representations.

\subsubsection{Multi-Proposal Module}\label{sec:multi-proposal}
Most previous action recognition datasets have only one action (or interaction) at any time, regardless of the number of people present. As a result, previous approaches for skeleton action recognition cannot handle the scenario where multiple people perform different actions simultaneously. When there are multiple people in the scene, they simply do a max over features extracted from each of them, and forward the resulting combined feature to output one action. Thus, they can only predict one action at a time. 

However, in our dataset, when there are multiple people in the scene, they are free to do any actions or interact with each other at any time. So there are many scenarios where multiple people are doing actions and interacting simultaneously. We tackle this problem with a multi-proposal module. Specifically, denote $N$ to be the number of people appearing at the same time. Instead of performing max-pooling over $N$ features, our multi-proposal module outputs $N+{N\choose 2}$ proposals from these $N$ features, corresponding to $N$ possible single-person actions and ${N\choose 2}$ possible interactions between each two people. Our multi-proposal module enables us to output multiple actions and interactions at the same time. Finally, we adopt a priority strategy to prioritize interactions over single person actions. For instance, if there are predictions for `pointing to something' (single person) and `pointing to someone' (interaction) at the same time, our final prediction would be `pointing to someone'.


\subsection{Multimodal End-to-end Training}
Since we want to train our model in an end-to-end manner, we can no longer use $\arg\max$ to extract 3D keypoint locations, as in past work on RF-based pose estimation~\cite{zhao2018rf}. Thus, we use a regressor to perform the function of the $\arg\max$ to extract the 3D locations of each keypoint. This makes the model differentiable and therefore the action label can also act as supervision on the skeleton prediction model.

Our end-to-end architecture uses 3D skeletons as an intermediate representation which enables us to leverage previous skeleton based action recognition datasets. We combine different modalities to train our model in the following manner: for wireless signal datasets, gradients back propagate through the whole model, and they are used to tune the parameters of both the skeleton prediction model and the action recognition model; for previous skeleton-based action recognition datasets, the gradients back propagate till the skeleton, and they are used to tune the parameters for the action recognition module. As shown in the experiments section, this multi-modality training significantly increases the data diversity and improves the performance of our model.
\begin{figure*}[t]
\begin{center}
\includegraphics[width=1.0\linewidth, height=0.6\linewidth]{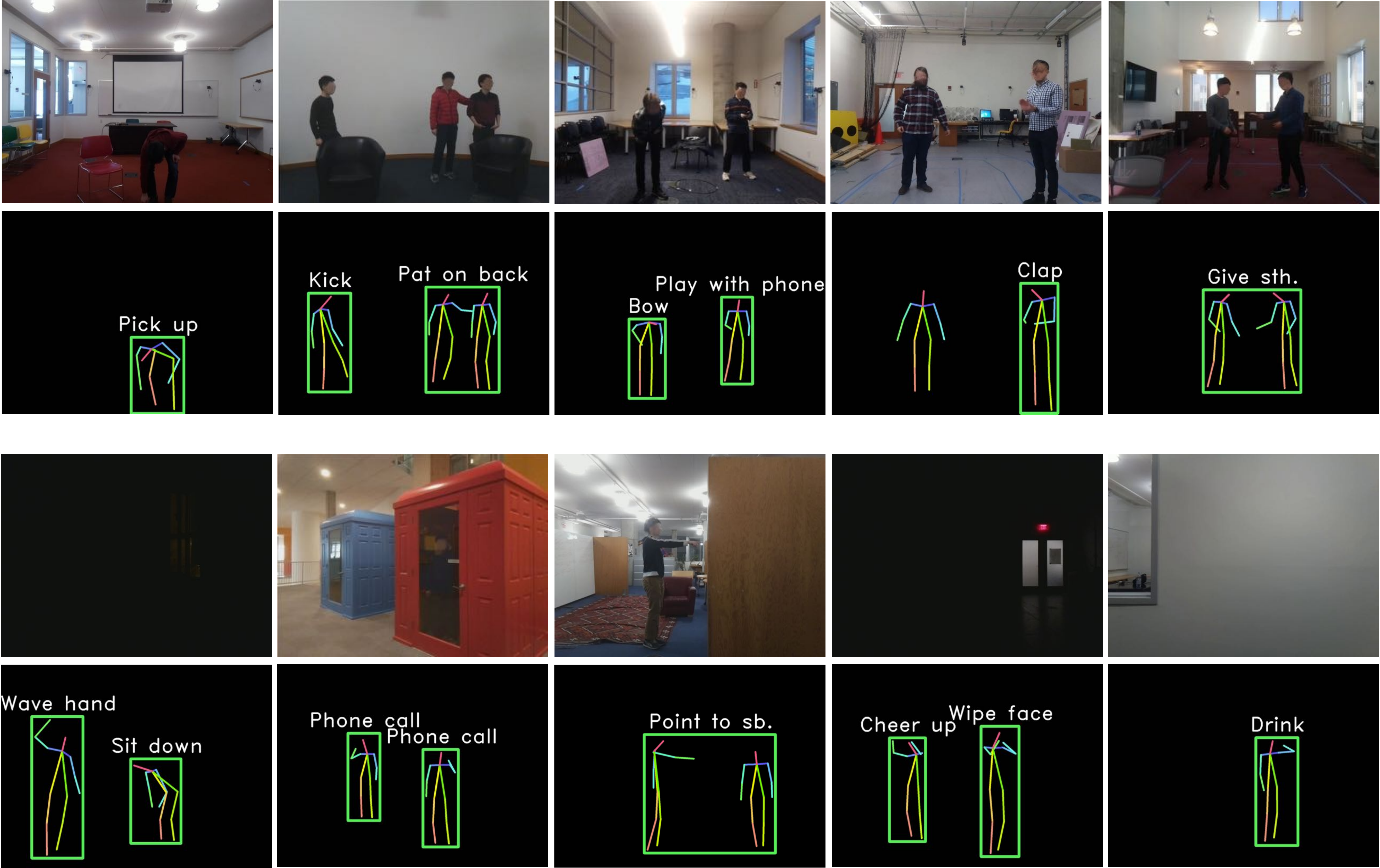}
\end{center}
\vspace{-10pt}
\caption{\footnotesize{Qualitative Results. The figure shows \name's output in various scenarios. The top two rows show our model's performance in visible scenes. The bottom two rows show our model's performance under partial/full occlusions and poor lighting conditions. The skeletons shown are the 2D projection of the intermediate 3D skeletons generated by our model.}}\label{fig:main}
\vspace{-10pt}
\end{figure*}

\section{Experiments}

\subsection{Dataset}
Since none of the available action detection datasets provide RF signals and the corresponding skeletons, we collect our own dataset which we refer to as RF Multi-Modality Dataset (RF-MMD). We use a radio device to collect RF signals, and a camera system with 10 different viewpoints to collect video frames. The radio device and the camera system are synchronized to within 10 ms. Appendix A includes a more detailed description of our data collection system.

We collected 25 hours of data with 30 volunteers in 10 different environments, including offices, lounges, hallways, corridors, lecture rooms, etc. 
%
We choose 35 actions (29 single actions and 6 interactions) from the PKU-MMD's action set~\cite{liu2017pku}. For every 10-min data, we ask up to 3 volunteers to perform different actions randomly from the above set. On average, each sample contains 1.54 volunteers, each volunteer performs 43 actions within 10 minutes, and each action takes 5.4 seconds. We use 20 hours of our dataset for training and 5 hours for testing. 

The dataset also contains 2 through-wall scenarios, where one is used for training and one for testing. As for these through-wall environments, we put cameras on each side of the wall so that the camera system can be calibrated with the radio device, and use those cameras which can see the person to label the actions. All test result on RF-MMD use only radio signals without vision-based input.

We extract 3D skeleton sequences using the multi-view camera system \cite{zhao2018rf}.  We first apply AlphaPose \cite{fang2017rmpe} to the videos collected by our camera system to extract multi-view 2D skeletons. Since there may be multiple people in the scene, we then associate the 2D skeletons from each view to get the multi-view 2D skeletons for each person. After that, since our camera system is calibrated, we can triangulate the 3D skeleton of each person. These 3D skeletons act as the supervision for the intermediate 3D skeletons generated by our model.

Finally, we leverage the PKU-MMD dataset~\cite{liu2017pku} to provide additional training examples.  The dataset allows for action detection and recognition. It contains almost 20,000 actions from 51 categories performed by 66 subjects. This dataset allows us to show how \name\ learns from vision-based examples.

\subsection{Setup}
\vskip 0.06in\noindent
\textbf{Metric.}
As common in the literature on video-based action detection  \cite{lin2017single,zhao2017temporal,Heilbron_2015_CVPR} and skeleton-based action detection \cite{liu2017pku,li2017skeleton,li2018co}, we evaluate the performance of our model using the mean average precision (mAP) at different intersection-over-union (IoU) thresholds $\theta$. We report our results on mAP at $\theta=0.1$ and $\theta=0.5$. 

\vskip 0.06in\noindent
\textbf{Ground Truth Labels.}
To perform end-to-end training of our proposed \name\ model, we need two types of ground truth labels: 3D human skeletons to supervise our intermediate representation, and action start-end time and category to supervise the output of our model. The 3D skeletons are triangulated using AlphaPose and the multi-view camera system described earlier. As for actions' duration and category, we manually segment and label the action of each person using the multi-view camera system.

\subsection{Qualitative Results}
Figure \ref{fig:main} shows qualitative results that illustrate the output of \name\ under a variety of scenarios.  The figure shows that \name\ correctly detects actions and interactions, even when different people perform different actions simultaneously, and can deal with occlusions and poor lighting conditions. Hence, it addresses multiple challenges for today's action recognition systems.

\subsection{Comparison of Different Models}

We compare the performance of \name\ to the state-of-the-art models for skeleton-based action recognition and RF-based action recognition. We use HCN as a representative of a top performant skeleton-based action detection system in computer vision. It currently achieves the best accuracy on this task.  We use Aryokee~\cite{tian2018rf} as a representative of the state-of-the-art in RF-based action recognition. To our knowledge, this is the only past RF-based action recognition system that performs action detection in addition to classification.\footnote{The original Aryokee code is for two classes. So we extended to support more classes.} All models are trained and tested on our RF action recognition dataset. Since HCN takes skeletons as input (as opposed to RF signals), we provide it with the intermediate skeletons generated by \name. This allows us to compare \name\ to HCN in terms of action recognition based on the same skeletons. 

\begin{table}[htbp]
  \small
    \centering
    \vspace{-5pt}
    \begin{tabular}{ccccc}
        \hline
        \multirow{3}{*}{Methods} & \multicolumn{2}{c}{Visible scenes} & \multicolumn{2}{c}{Through-wall}\\ \cmidrule(lr){2-3} \cmidrule(lr){4-5}
        & \multicolumn{2}{c}{mAP} & \multicolumn{2}{c}{mAP}\\ 
        & $\theta$=0.1 & $\theta$=0.5 & $\theta$=0.1 & $\theta$=0.5 \\
        \hline
        \name\ & \textbf{90.1} & \textbf{87.8} & \textbf{86.5} & \textbf{83.0}\\
        HCN \cite{li2018co} & 82.5 & 80.1 & 78.5 & 75.9  \\
        Aryokee \cite{tian2018rf} & 78.3 & 75.3 & 72.9 & 70.2 \\
        \hline
    \end{tabular}
  \vspace{-10pt}
    \caption{\footnotesize{Model Comparison on RF-MMD dataset. The table shows mAP in visible and through-wall scenarios under different IoU threshold $\theta$. Since HCN operates on skeletons, and for fair comparison, we provide it with the RF-based skeletons generated by \name.}    }
  \vspace{-10pt}
  \label{tab:ap_main}
\end{table}

 Table \ref{tab:ap_main} shows the results for testing on visible scenes and through-wall scenarios, with wireless signals as the input.  As shown in the table, \name\ outperforms HCN in both testing conditions. This shows the effectiveness of our proposed modules. Further, we can also see that \name\ outperforms Aryokee by a large margin on both visible and through-wall scenarios. This shows that the additional supervision from the skeletons, as well as \name\ neural network design, are important for delivery of accurate performance using RF data.

\subsection{Comparison of Different Modalities}
Next, we investigate the performance of \name\ when operating on RF-based skeletons versus vision-based skeletons. 
We train \name\ on the training set, as before. However, when performing inference, we either provide it with the input RF signal from the test set, or we provide it with the visible ground truth skeletons obtained using our camera system. 
Table \ref{tab:ap_main_vision} shows the results for different input modalities. The table shows that for visible scenes, operating on the ground truth skeletons from the camera system leads to only few percent improvements in accuracy. This is expected since the RF-skeletons are trained with the vision-based skeleton as ground truth. Further, 
as we described in our experimental setting, the camera-based system uses 10 viewpoints to estimate 3D skeletons while only one wireless device is used for action recognition based on RF. This result demonstrates that RF-based action recognition can achieve a performance close to a carefully calibrated camera system with 10 viewpoints. The system continues to work well in through-wall scenarios though the accuracy is few percents lower due to the signal experiencing some attenuation as it traverses walls. 

\begin{table}[htbp]
  \small
      \vspace{-5pt}
    \centering
    \begin{tabular}{ccccc}
        \hline
        \multirow{3}{*}{Method / Skeletons } & \multicolumn{2}{c}{Visible scenes} & \multicolumn{2}{c}{Through-wall}\\ \cmidrule(lr){2-3} \cmidrule(lr){4-5}
        & \multicolumn{2}{c}{mAP} & \multicolumn{2}{c}{mAP}\\ 
        & $\theta$=0.1 & $\theta$=0.5 &  $\theta$=0.1 & $\theta$=0.5 \\
        \hline
        \name\ / RF-MMD & 90.1 & 87.8 & 86.5 & 83.0\\
        \name\ / G.T. Skeleton & 93.2 & 90.5 & - & - \\
        \hline
    \end{tabular}
  \vspace{-10pt}
    \caption{\footnotesize{\name's Performance (mAP) with RF-Based Skeletons (RF-MMD) and Vision-Based Skeleton (G.T. Skeleton) under different IoU threshold $\theta$.}}
  \vspace{-10pt}
  \label{tab:ap_main_vision}
\end{table}

\subsection{Action Detection}

\begin{figure}[htbp]
\centering
\includegraphics[width=\linewidth]{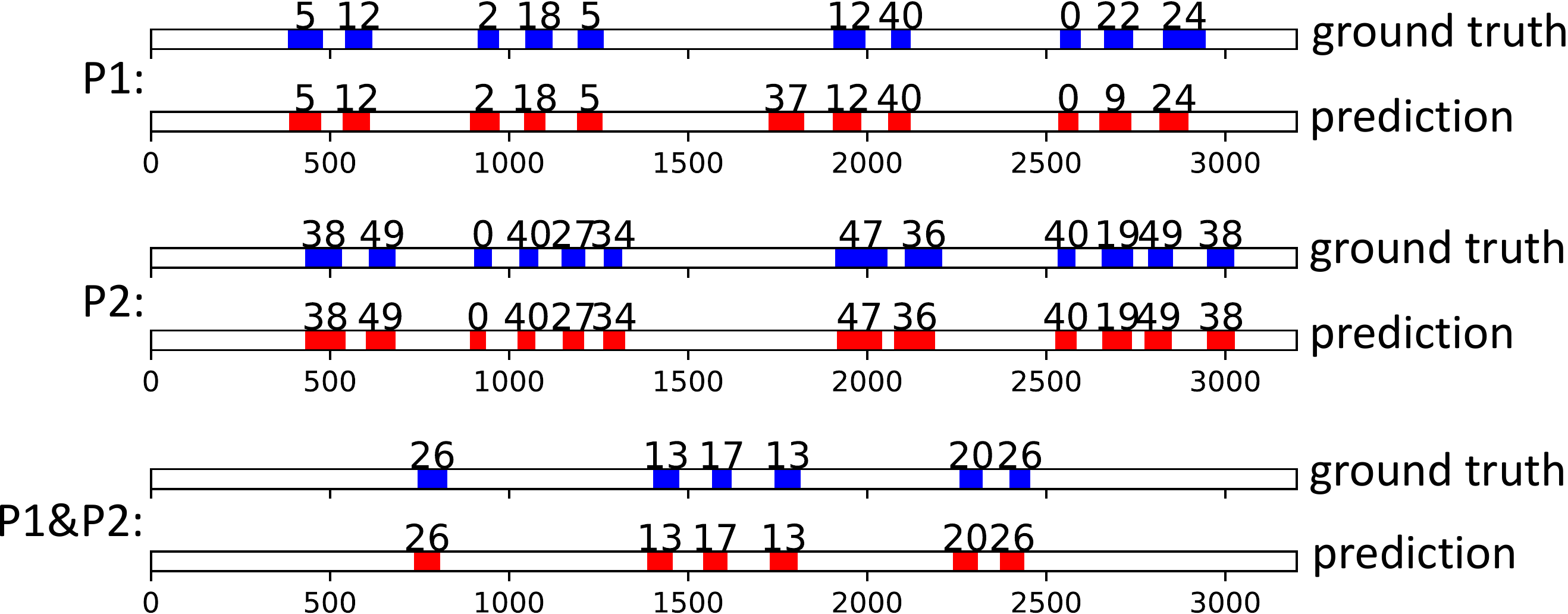}
  \vspace{-10pt}
\caption{\footnotesize{Example of action detection results on the test set, where two people are doing actions as well as interacting with each other. Ground truth action segments are drawn in blue, while detected segments using our model are drawn in red. The horizontal axis refers to the frame number.}}    \label{fig:detection-result}
  \vspace{-10pt}
\end{figure}

In Figure \ref{fig:detection-result}, we show a representative example of our action detection results on the test set. Two people are enrolled in this experiment. They sometimes do actions independently, or interact with each other. The first row shows the action duration for the first person, the second row shows the action duration of the second person, and the third row shows the interactions between them. Our model can detect both the actions of each person and the interactions between them with high accuracy. This clearly demonstrates that our multi-proposal module has good performance in scenarios where multiple people are independently performing some actions or interacting with each other. 

\subsection{Ablation Study}
We also conduct extensive ablation studies to verify the effectiveness of each key component of our proposed approach. For simplicity, the following experiments are conducted on the visible scenes in RF-MMD and mAP are calculated under 0.5 IoU threshold.
\vskip  0.06in \noindent
\textbf{Attention Module. }
We evaluate the effectiveness of our proposed spatial-temporal attention module in Table \ref{tab:ap_attention}. We show action detection performance with or without our attention module on both RF-MMD and PKU-MMD. The results show that our attention is useful for both datasets, but is especially useful when operating on RF-MMD. This is because skeletons predicted from RF signals can have inaccurate joints. We also conduct experiments on the NTU-RGB+D ~\cite{shahroudy2016ntu} dataset. Unlike PKU-MMD and RF-MMD which allow for action detection, this dataset is valid only for action classification. 
The table shows that our attention module is useful in this case too. 

\begin{table}[htbp]
  \small
    \centering
    \vspace{-5pt}
     \begin{tabular}{ccc}
         \hline
        Datasets (Metric)& \name\ & \name\ w/o Attention \\
         \hline
         RF-MMD (mAP) & \textbf{87.8} & 80.1 \\
        PKU-MMD (mAP) & \textbf{92.9}/~\textbf{94.4} & 92.6/~94.2 \\
        NTU-RGB+D (Acc) & \textbf{86.8}/~\textbf{91.6} & 86.5/~91.1 \\
         \hline
     \end{tabular}
  \vspace{-10pt}
      \caption{\footnotesize{Performance of \name\ on Different Datasets With and Without Attention. For PKU-MMD and NTU-RGB+D (cross subject /cross view), we test the action recognition network (without the skeleton generation network). Tests on RF-MMD are across subjects and environments. }}
        \vspace{-10pt}
  \label{tab:ap_attention}
\end{table}

\vskip  0.06in \noindent
\textbf{Multi-Proposal Module.}
We propose a multi-proposal module to enable multiple action prediction at the same time. We evaluate our model's performance with or without the multi-proposal module. As shown in Table \ref{tab:ap_multi_proposal}, the added multi-proposal module significantly increase the performance. This is because our dataset includes a lot of instances when people are performing different actions at the same time. Our model will get very poor accuracy at these scenarios with single-proposal, while with multi-proposal our model can achieve much higher performance.

\begin{table}[htbp]
  \small
    \centering
    \begin{tabular}{cc}
        \hline
        Methods & RF-MMD \\
        \hline
        Multi-Proposal & 87.8 \\
        Single-Proposal & 65.5 \\
        \hline
    \end{tabular}
  \vspace{-10pt}
      \caption{\footnotesize{Benefits of Multi-proposal Module. The table shows adding multi-proposal module largely improves the performance on RF-MMD}}
        \vspace{-5pt}
  \label{tab:ap_multi_proposal}
\end{table}

\vskip  0.06in \noindent
\textbf{Multimodal Training.}
As explained earlier, the use of skeletons as an intermediate representation allows the model to learn from both RF datasets and vision-based skeleton datasets. To illustrate this advantage, we perform multimodal training by adding PKU-MMD's training set into the training of our RF-Action model. More specifically, we use our dataset to train the whole \name\ end-to-end model, and use the PKU-MMD dataset to train \name's activity detection model. These two datasets are used alternatively during training. As shown in Table \ref{tab:ap_joint}, comparing the detection results with the model trained on either dataset separately, we find multimodal training can increase the model performance since it introduces more data for training and thus can get better generalization ability. 

\begin{table}[htbp]
  \small
    \centering
    \vspace{-5pt}
    \begin{tabular}{ccc}
        \hline
        Training set $\backslash$ Test set & RF-MMD & PKU-MMD \\
        \hline
        RF-MMD+PKU-MMD & \textbf{87.8} & \textbf{93.3}/~\textbf{94.9} \\
        RF-MMD & 83.3 & 60.1/~60.4 \\
        PKU-MMD & 77.5 & 92.9/~94.4 \\
        \hline
    \end{tabular}
  \vspace{-10pt}
    \caption{\footnotesize{Benefits of Multimodal Training. The table shows that adding PKU-MMD to the training set significantly improves the performance on RF-MMD. The mAP of RF-MMD+PKU-MMD on RF-MMD are achieved using the cross-subject training set of PKU-MMD. Using only RF-MMD for training has a poor performance on PKU-MMD because the action set of RF-MMD is only a subset of PKU-MMD's action set.}}
  \vspace{-5pt}
  \label{tab:ap_joint}
\end{table}

\vskip  0.06in \noindent
\textbf{End-to-End Model.}
\name\ uses an end-to-end model where the loss of action recognition is back propagated through the skeleton generation network. Here we show that such an end-to-end approach improves the skeleton itself. Table \ref{tab:ap_e2e} reports the average error in skeleton joint location, for two systems: our end-to-end model and an alternative model where the skeleton is learned separately from the action --i.e., the action loss is not propagated through the skeleton generation network. The table shows that the end-to-end model not only improves the performance of the action detection task, but also reduces the errors in estimating the location of joints in RF-based skeletons. This is because the action detection loss provides regularization for 3D skeletons generated from RF signals. 

\begin{table}[htbp]
  \small
    \centering
    \begin{tabular}{ccc}
        \hline
        Methods & mAP & Skeleton Err. (cm) \\
        \hline
        end-to-end & 87.8 & 3.4 \\
        separate & 84.3 & 3.8 \\
        \hline
    \end{tabular}
  \vspace{-10pt}
    \caption{\footnotesize{mAP and intermediate 3D skeleton error on testing data with and without end-to-end training.}}
  \vspace{-5pt}
  \label{tab:ap_e2e}
\end{table}


\vspace{-10pt}
\section{Conclusion}
This paper presents the first model for skeleton-based action recognition using radio signals, and demonstrates that such model can recognize actions and interactions through walls and in extremely poor lighting conditions. 
The new model enables action recognition in settings where cameras are hard to use either because of privacy concerns or poor visibility.  Hence, it can bring action recognition to people's homes and allow for its integration in smart home systems. 

\clearpage
{\small
\bibliographystyle{ieee_fullname}
\bibliography{main}
}

\end{document}